\documentclass[conference]{IEEEtran}
\usepackage{times}

\usepackage[numbers]{natbib}
\usepackage{multicol}
\usepackage[bookmarks=true]{hyperref}

\usepackage{hyperref}       
\usepackage{booktabs}       
\usepackage{amsfonts}       
\usepackage{nicefrac}       
\usepackage{microtype}      
\usepackage{graphicx}
\usepackage{amssymb,amsmath,mathrsfs}
\usepackage{comment,multirow,graphicx,bbm,subcaption,caption,mathtools}
\usepackage{mathabx,algorithmic,algorithm}
\usepackage{balance}
\usepackage[normalem]{ulem}  

\graphicspath{ {figs/} }

\usepackage{xcolor}

\newcommand{\mypara}[1]{\vspace{0.7em}\noindent\textbf{#1}}

\newcommand{\CLASSINPUTtoptextmargin}{19mm}
\newcommand{\CLASSINPUTbottomtextmargin}{18mm}
\newcommand{\CLASSINPUTinnersidemargin}{19mm}
\newcommand{\CLASSINPUToutersidemargin}{19mm}

\begin{document}

\title{\vspace{6mm}Extended Tactile Perception: Vibration Sensing through Tools and Grasped Objects}

\author{\IEEEauthorblockN{Tasbolat Taunyazov\IEEEauthorrefmark{1}, Luar Shui Song\IEEEauthorrefmark{1}, Eugene Lim\IEEEauthorrefmark{1}, Hian Hian See\IEEEauthorrefmark{2},\\David Lee\IEEEauthorrefmark{2}\IEEEauthorrefmark{3}, Benjamin C.K. Tee\IEEEauthorrefmark{2}\IEEEauthorrefmark{3}, and Harold Soh\IEEEauthorrefmark{1}}
\IEEEauthorblockA{\IEEEauthorrefmark{1}\textit{Dept. of Computer Science, National University of Singapore}}
\IEEEauthorblockA{\IEEEauthorrefmark{2}\textit{Dept. of Materials Science and Engineering, National University of Singapore}\\
\IEEEauthorblockA{\IEEEauthorrefmark{3}\textit{Institute for Health Technology and Innovation, National University of Singapore}}
Email: tasbolat@comp.nus.edu.sg, luarss@comp.nus.edu.sg, elimwj@nus.edu.sg  mseshh@nus.edu.sg,\\david.leekh@nus.edu.sg, benjamin.tee@nus.edu.sg, harold@comp.nus.edu.sg
}
}

\maketitle

\begin{abstract}
Humans display the remarkable ability to sense the world through tools and other held objects. For example, we are able to pinpoint impact locations on a held rod and tell apart different textures using a rigid probe. In this work, we consider how we can enable robots to have a similar capacity, i.e., to embody tools and extend perception using standard grasped objects. We propose that vibro-tactile sensing using dynamic tactile sensors on the robot fingers, along with machine learning models, enables robots to decipher contact information that is transmitted as vibrations along rigid objects. This paper reports on extensive experiments using the BioTac micro-vibration sensor and a new event dynamic sensor, the NUSkin, capable of multi-taxel sensing at 4~kHz. We demonstrate that fine localization on a held rod is possible using our approach (with errors less than 1 cm on a 20 cm rod). Next, we show that vibro-tactile perception can lead to reasonable grasp stability prediction during object handover, and accurate food identification using a standard fork. We find that multi-taxel vibro-tactile sensing at a sufficiently high sampling rate led to the best performance across the various tasks and objects. Taken together, our results provide both evidence and guidelines for using vibro-tactile perception to extend tactile perception, which we believe will lead to enhanced competency with tools and better physical human-robot interaction.
\end{abstract}

\IEEEpeerreviewmaketitle

\section{Introduction}
\label{sec:intro}

Our proficiency with our hands is crucially supported by our sense of touch. Mechanoreceptors in our skin enable us to finely perceive tactile information, which is invaluable when directly manipulating objects and when using tools. Indeed, humans can accurately localize contact not only on our skin, but also on grasped objects~\cite{Miller2018}, and are able to discriminate textures through rigid links~\cite{Klatzky1999}. 
Unlike a majority of tactile perception, which has focused on determining properties of grasped objects, we are inspired by our remarkable ability to sense the world \emph{through} tools and other held items. 

In this work, we seek to extend robot tactile perception by using \emph{standard} objects (e.g., a stick or a fork) and \emph{without} the use of mounted accelerometers. 
Prior work has demonstrated that contact \emph{events} --- whether/when a contact occurred on a held object --- can be reliably detected with existing sensors. For example, tactile signals have been used to trigger motions during human-robot handover~\cite{Yamaguchi2019} and to detect when a held soda-can has been placed on a table~\cite{Romano2011}. Here, we go beyond existing research and ask: \emph{can we decipher the properties of such contacts}? Where on a held tool's surface did the contact occur? What type of object did the tool make contact with? 

\begin{figure}
\centering
\includegraphics[width=0.95\columnwidth]{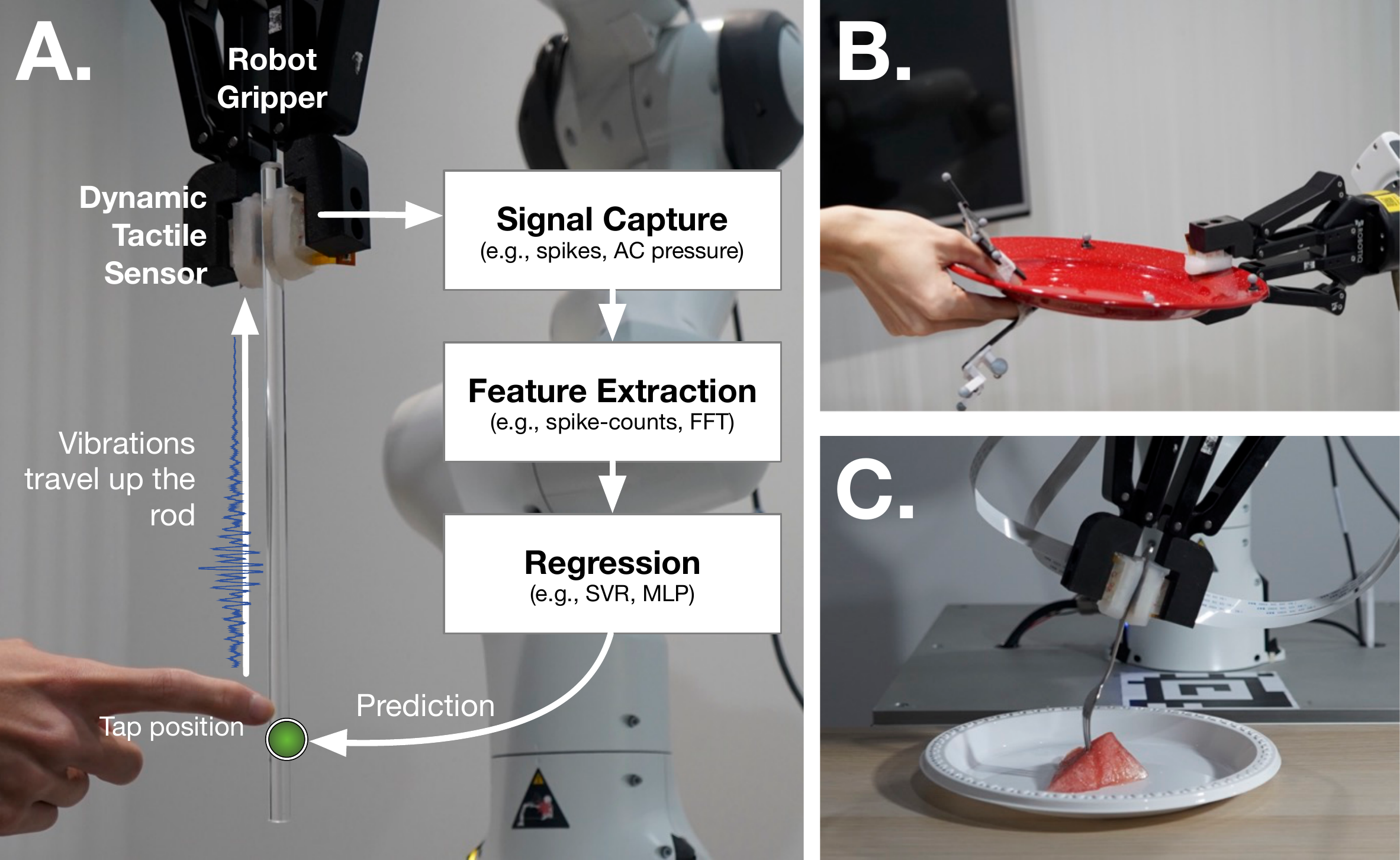}
\caption{\small We study Extended Tactile Perception --- our goal is to enable robots to extend their  tactile perception through \emph{standard} objects such as tools. (\textbf{A}) We show that robots are able to accurately localize taps on an acrylic rod using fast vibro-tactile sensing and machine learning. Vibrations caused by the tap travel up the rod where they are picked up by a dynamic tactile sensor (the NUSkin in this image). The signal is captured and mapped into a tap position using simple models and learned features. We provide results on two additional tasks: (\textbf{B}) grasp stability classification during object handover and (\textbf{C}) food classification through a fork.}
\label{fig:summary}	
\end{figure}

Addressing this challenge is important for enabling competency with tools --- sensing via tools can help robots determine the properties of objects that are out of reach, or that are too difficult or dangerous to interact with directly. 
In humans, recent evidence suggests vibration sensing via fast-adapting (FA) somatic sensory receptors and neural processing of ``vibratory motifs'' underlies our capacity to embody tools and extend perception beyond our body~\cite{Miller2018}.
Likewise, we hypothesize that extended tactile perception (beyond the boundary of the robot) may be achieved using a combination of high-frequency vibro-tactile sensing via artificial skin on the robot and processing via statistical learning or neural models (Fig. \ref{fig:summary}).

A key feature of our work is that we only use compact tactile sensors on the robot gripper, rather than specially-crafted sensorized tools (e.g., \cite{Bhattacharjee2019,Romano2014,Strese2017}). 
We propose a system that senses and learns to decode contact information that is transmitted as vibrations (or movements) along standard rigid objects, such as a rod/fork/spoon or a handled item such as a plate. Accurately detecting these vibrations requires a sensitive dynamic tactile sensor with high frequency response. In this work, we experiment with the BioTac, specifically its hydrophone pressure sensor, which has a frequency response that exceeds human mechanoreceptors. However, the BioTac currently has one such dynamic sensor. We contribute NUSkin; a 40-taxel event sensor based on NeuTouch~\cite{Taunyazov2020} that has been augmented to have a higher 4~kHz sampling rate. Unlike the BioTac, the NUSkin taxels can fire asynchronously --- each taxel generates positive or negative spikes depending on changes in the contact pressure. 

To interpret the vibro-tactile signals, we adopt a learning approach. 
However, with high sampling rates, the collected data comprise long sequences that can be difficult to learn from. This problem is exacerbated in data constrained settings such as tactile sensing where data collection is relatively costly. 
Moreover, the event sensor only outputs sparse spike trains (not pressure values) that can be difficult to handle in standard machine learning methods. We find that recent neural models achieved the best performance in our tests, but simple (binned/smoothed) spike-count and FFT features together with ``classical'' statistical learning methods  also worked well. 

Experiments show that by enabling robots to quickly perceive and make sense of vibratory features, they are also capable of accurately localizing contact on a grasped rod --- on a 20~cm rod, the localization error was only $\approx${1~cm}. Further experiments show that our approach can be applied to tasks such as stable-grasp detection during human-robot handover (where static force detection is usually applied), and food classification for robot feeding. Taken together, our results indicate that with vibro-tactile perception, a robot can extend its capacity to sense the world, akin to having ``virtual sensors'' on the held objects. 
 
To summarize, our key contributions are:
\begin{itemize}
	\item A novel method of vibro-tactile perception that enables robots to extend sensing through everyday objects;
	\item The NUSkin, a modification of a recently proposed event tactile sensor capable of higher sampling rates;
	\item Experimental results that validate the extended tactile perception system on a real-world robot, along with findings that can impact future designs.
\end{itemize}
In addition, we have made our code and datasets publicly available\footnote{\url{https://github.com/clear-nus/ext-sense}.}; they comprise the tactile signals captured during our experiments and ground-truth annotations, which we hope will  facilitate future work on extended tactile perception. We have focused primarily on vibrations in this work, and a plausible next step is examine multiple sensors in multi-modal framework. More broadly, we believe this paper brings us closer towards general tool embodiment for robotics.

\section{Background and Related Work}
\label{sec:background}

Tactile perception in robotics is a broad research area that spans  sensing, learning, and control. Here, we give a brief overview of related work and refer interested readers to survey articles~\cite{Luo2017,Yamaguchi2019,Silvera-Tawil2015,Kappassov2015} for more comprehensive information.  The goal of our study is to determine the properties of interactions that occur on a held/touched object. On the whole, prior work has focused on perceiving properties of objects that are in \emph{direct} contact with the sensor. A notable exception is closely-related work~\cite{molchanov2016contact}, which also used the BioTac with machine learning methods to determine contact locations on a variety of objects. Our work differs in several respects; we focus on dynamic tactile sensing and contribute the neuromorphic NUSkin sensor, novel analysis into the speed and number of taxels required, and experiments on additional tasks beyond contact localization.

\mypara{Dynamic Tactile Sensors.} There has been significant progress on tactile sensing and a range of sensors --- from piezoelectric to optical~\cite{yuan2017gelsight} --- have been developed over the years~\cite{Kappassov2015,Yamaguchi2019}. 
Here, we focus on \emph{dynamic tactile sensing}, i.e., the measurement of rapidly changing tactile information. Examples of dynamic tactile sensors include accelerometers~\cite{Cutkosky2014} and robot whiskers~\cite{Solomon2006}. The latter is notable since the whiskers themselves do not possess tactile sensing elements. Rather, deformations of the whiskers (sensed via the whisker's torque and angle at the base) can be used to sense the presence and shape of objects. 

Dynamic tactile sensing has also been performed using attached sensors or specially-designed tools. For example, notable work showed that vibrations (captured using an accelerometer or contact mic mounted on a container) obtained by shaking a vessel can reveal the type and approximate number of items within~\cite{Chen2017}. Prior work has devised haptic recording tools that employ accelerometers, microphones, and force sensors~\cite{Romano2014,Strese2017} to accurately discriminate textures. In recent work~\cite{Bhattacharjee2019}, a dinner fork instrumented with a 6-axis force/torque sensor (the ``Forque'') was used to perform food classification.

In this paper, we focus on a more general setting where dynamic tactile sensing is performed using sensors on the robot fingers. We work with two modern compact sensors: the SynTouch BioTac and the NUSkin, a new variant of the NeuTouch~\cite{Taunyazov2020}. In addition to impedance sensing electrodes, the BioTac has a hydro-acoustic pressure sensor that senses vibrations that propagate through the skin and incompressible conductive fluid in the finger~\cite{Fishel2008}. The sensor is capable of detecting micro-vibrations (frequency response up to 1040~Hz) and can discriminate textures better than humans~\cite{Fishel2012}. A multi-taxel alternative is the NeuTouch, which uses a piezoresistive graphene transducer with 39 taxels. 
Unlike the BioTac, the NeuTouch transmits spikes, which are generated by changes in pressure. These two sensors operate differently and comparing them in our experiments allowed us to examine different sensing modes, the role of multiple taxels, and type of transmitted data. In this work, we developed a modified version of the NeuTouch with a higher sampling rate of 4kHz (described in Sec. \ref{sec:tactilesensor}).
Our methodology could be applied to other sensors that are fast and sensitive enough to detect vibrations on the grasped object.

\mypara{Features for tactile-sensing.} 
To make sense of vibration and tactile signals, prior work has used a variety of features --- from handcrafted~\cite{Zhang2016,Pezzementi2011,Strese2017} and statistical descriptors~\cite{Soh2012,Soh2014} to features learned via dictionary methods~\cite{Schneider2009,Madry2014,Luo2014,Chen2017,Liu2018} and neural networks~\cite{Luo2017,Hoffmann2014,molchanov2016contact}. In addition to raw temporal data, we work with a variety of features: Fast Fourier Transform (FFT)~\cite{brigham67} features, learnt neural representations via autoencoding~\cite{kramer1991nonlinear} and the recently proposed Event Spike Tensor~\cite{gehrig2019end}.

\mypara{Applications of Vibro-Tactile Signals.} Within the context of robotics, tactile sensing is widely recognized as important for dexterous manipulation~\cite{Billard2019} and a variety of robot tasks (e.g., grasping~\cite{Romano2011,Chebotar2016}, slip-detection~\cite{Chen2018,Taunyazov2020},  handovers~\cite{GomezEguiluz2019,Hendrich2014}, and robot feeding~\cite{Bhattacharjee2019}). However, prior work has focused on identifying contact events or properties of the grasped object (e.g., its texture~\cite{Taunyazov2019}, identity~\cite{Bhattacharjee2019,Soh2012,Soh2014}, or even its contents~\cite{Chen2017,Matl2019}). In this work, we focus on decoding vibratory signals that are transferred \emph{through} the held/touched object. Vibration analysis has long been applied in numerous fields, e.g., in the monitoring and diagnosis of machines, but the precise problem in this paper --- the interpretation of vibrations through grasped tools --- remains relatively unexplored in robotics.

\section{Problem Statement: Extended Tactile Perception}
\label{sec:problem}

In this work, our overarching goal is to accurately perceive the properties of contacts on a held/touched rigid object; a problem that we call \emph{extended tactile perception}. 

Consider a simple straight rod that has touched another object, say a piece of fruit. Information about point(s) of contact and properties of the fruit (e.g., softness, roughness) is transmitted along the rod as mechanical transients~\cite{Cutkosky1993,Miller2018}. The vibrations arising from an impulse along the rod can be modeled using partial differential equations (PDEs) from Euler-Bernoulli Beam theory~\cite{Timoshenko1953}. However, finding solutions to the PDE is difficult in general and requires precise knowledge of the rod's material properties. Moreover, complex interactions (e.g., arising from further manipulation) or with non-uniformly shaped tools (e.g., a fork) are difficult to model analytically.

To sidestep these issues, we adopt a learning-based approach. A tactile sensor produces a stream or sequence of data $\mathbf{x} = [(\mathbf{x}_1, t_1), (\mathbf{x}_2, t_2), \dots, (\mathbf{x}_{T}, t_{T})] \in \mathcal{X}$, where each $\mathbf{x}_j \in \mathbb{R}^d$ is the observed tactile data and $t_k$ is the time it was observed. In general, the timings $t_k$ may be irregular and each sequence $\mathbf{x}^{(i)}$ can have a different length $T^{(i)}$. Our task is to associate each observed signal $\mathbf{x}$ with a contact property $\mathbf{y} \in \mathcal{Y}$, e.g., the contact location. Given a dataset $\mathcal{D} = \{ (\mathbf{x}^{(i)}, \mathbf{y}^{(i)}) \}_{i=1}^N$ of $N$ tuples, we seek to learn a parameterized function $f_\theta:\mathcal{X}\rightarrow \mathcal{Y}$ that well captures this association. 

For our perception system to work well, there are several major challenges to address. The sensor must be sufficiently sensitive with a frequency response high enough to detect vibrations. The Pacinian and Meissner's corpuscles in human skin detect vibration frequencies as high as 400Hz~\cite{Johansson2009}. To achieve similar (or better) performance, sampling should be performed at least above the Nyquist rate of 800Hz\footnote{The Nyquist rate is a sufficient condition and sub-Nyquist rate sampling may be possible for sparse or compressible signals.}. But at these high rates, the tactile data collected are long time series that may be difficult to interpret. Key questions arise about what methods and features may be suitable. It is the goal of this paper to provide a crucial first-step towards answering these questions.

\section{Event-based Tactile Sensor}
\label{sec:tactilesensor}

\begin{figure}
\centering
\includegraphics[width=0.4\columnwidth]{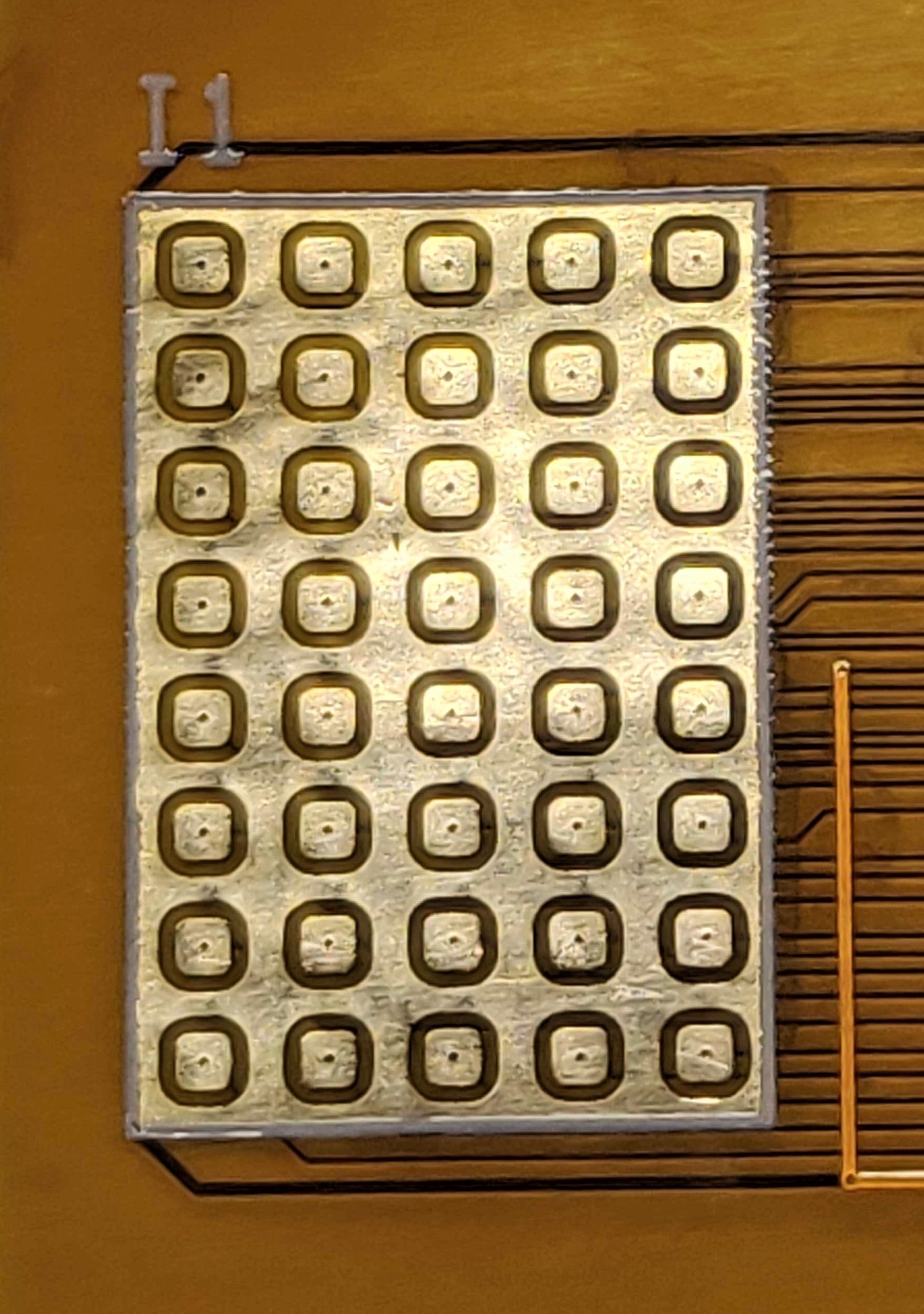}
\caption{Spatial distribution of the 40 taxels on NUSkin.}
\label{fig:electrode_layer}	
\end{figure}

Motivated by the requirements of vibration sensing, we contribute the NUSkin, a variant of the recently proposed NeuTouch event tactile sensor~\cite{Taunyazov2020}. 

We refer readers to \cite{Taunyazov2020} for details, but in summary, the NeuTouch is a neuromorphic tactile sensor that is similar to a human fingertip in size (37 x 21 x 13 mm), with Ecoflex-0030 as ``skin'' and 3D-printed ``bone'' as a base. Sensing is performed using an electrode array of 39 radially-arranged taxels and a graphene-based piezoresistive thin film for vibro-tactile sensing. The graphene-based pressure transducer can capture vibrational frequencies up to 1500~Hz~\cite{yao2020environment}, similar  to the BioTac. This highly-sensitive and low-hysteresis transducer helps to reduce the sensors’ response time and enables the capture of the high frequency vibratory signals. Transmission is performed using the Asynchronously Coded Electronic Skin (ACES)~\cite{Lee2019}, which enables each taxel to asynchronously transmit pressure changes in the form of positive and negative spikes (without time synchronization). This is achieved by encoding the taxels with unique electrical pulse signatures that are designed to overlap robustly for deconvolution. 

The NUSkin revises the NeuTouch in three specific ways:
\begin{itemize}
    \item \emph{Higher 4~kHz Sampling Rate.} The sampling rate was increased from 1~kHz up to 4~kHz, which better accommodates the frequency response of the graphene-based pressure transducer. This is made possible by reducing the pulse signature duration from 1 ms to 0.25 ms; 
    \item \emph{Regular taxel configuration.} The NUSkin has an electrode array of 40 taxels (v.s. 39 on the NeuTouch) organized in a lattice structure --- this structure is simple and allows the sensors to produce matrix-structured data, which facilities the application of deep learning methods; 
    \item \emph{Softer Skin.} The material to emulate the ``skin'' was replaced with Ecoflex 00-10 that has a lower Shore A Hardness value. This softer material helps to further amplify the stimuli exerted on the sensors and results in larger deformation during contact. 
    The former allows more spatial-temporal features to be collected~\cite{callier2019neural} and the latter provides a more stable grasp of tools and objects~\cite{khurshid2011robotic}, which aids in the transmission of vibratory signals to the transducer.
\end{itemize}

\section{Experiment Setup: Tasks and Robots}
\label{sec:experiments}

Our experiments were designed to test our primary hypothesis that fast vibro-tactile perception enables sensing through held objects.
We seek to answer fundamental questions related to vibration-based extended tactile perception: \emph{Can robots sense the world through tools by interpreting vibrations? How fast does sensing need to be performed? Is there a strong reason to use multiple taxels?} 

To that end, we conducted a systematic study using three different tasks and objects. In the remainder of this section, we give details on the tasks, robot setup, and methods used in our experiments. 
\begin{figure}
\includegraphics[width=\columnwidth]{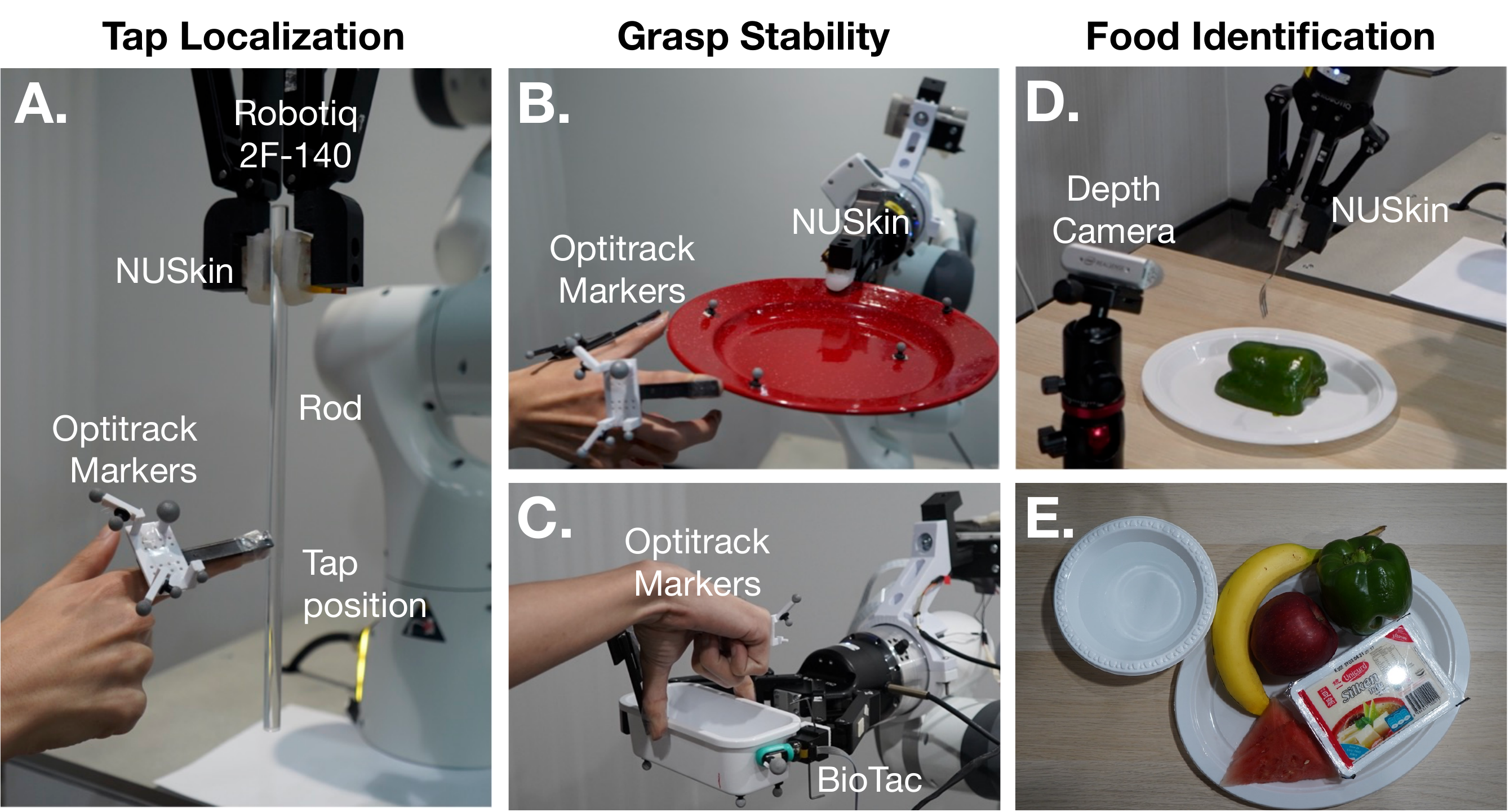}
\caption{\small Our experiments involved 3 different tasks: (\textbf{A}) Tap localization on a rod, (\textbf{B} and \textbf{C}) Grasp stability prediction on the a plate, a 30cm rod, and a box. In the shown images, the human has touched the plate, but has not yet achieved a stable grasp. In comparison, the box has been adequately grasped by the human. (\textbf{D} and \textbf{E}) Food identification through a standard fork with six different items (and a control class with nothing on the plate). Please see main text for details.}
\label{fig:experiments}	
\end{figure}
As a quick overview, our tasks along with their corresponding experimental setups are shown in Fig. \ref{fig:experiments}.

\begin{table*}[!ht]
\centering
\caption{\small MAE scores for the Tap Localization Task.}
\begin{tabular}{l|l|cc|cc|cc}
\hline \hline
\multirow{2}{*}{\textbf{Methods}}    & \multirow{2}{*}{\textbf{Features}} & \multicolumn{2}{c|}{\textbf{20cm}}                                                    & \multicolumn{2}{c|}{\textbf{30cm}}                                                    & \multicolumn{2}{c}{\textbf{50cm}}                                                    \\ \cline{3-8} 
                                     &                                    & \multicolumn{1}{c}{\textbf{BioTac (PAC)}} & \multicolumn{1}{c|}{\textbf{NUSkin (2F)}} & \multicolumn{1}{c}{\textbf{BioTac (PAC)}} & \multicolumn{1}{c|}{\textbf{NUSkin (2F)}} & \multicolumn{1}{c}{\textbf{BioTac (PAC)}} & \multicolumn{1}{c}{\textbf{NUSkin (2F)}} \\ \hline
\multirow{4}{*}{\textbf{SVR Linear}} & \textbf{Baseline}                  & 2.5492 ± 0.1873                           & 1.7102 ± 0.0913                           & 3.8219 ± 0.2584                           & 3.2682 ± 0.1861                           & 6.3491 ± 0.5526                           & 4.2323 ± 0.4525                          \\
                                     & \textbf{FFT}                       & 2.2397 ± 0.0821                           & 1.3864 ± 0.0448                           & 3.4202 ± 0.2153                           & 2.7135 ± 0.1811                           & 6.5452 ± 0.3566                           & 3.8361 ± 0.5234                          \\
                                     & \textbf{Autoencoder}               & 2.6015 ± 0.2132                           & 2.1976 ± 0.1058                           & 4.3844 ± 0.1394                           & 4.0754 ± 0.3331                           & 7.4146 ± 0.4647                           & 6.0886 ± 0.4546                          \\
                                     & \textbf{EST}                       & -                                         & 1.5260 ± 0.1176                           & -                                         & 2.7310 ± 0.0862                           & -                                         & 3.8520 ± 0.4105                          \\ \hline
\multirow{4}{*}{\textbf{SVR RBF}}    & \textbf{Baseline}                  & 2.0729 ± 0.1506                           & 1.5411 ± 0.1029                           & 3.0808 ± 0.1454                           & 2.8592 ± 0.1871                           & 5.2913 ± 0.4026                           & 3.5130 ± 0.4029                          \\
                                     & \textbf{FFT}                       & 1.9021 ± 0.0516                           & 1.2053 ± 0.0376                           & 2.7636 ± 0.0370                           & 2.3250 ± 0.1395                           & 6.2178 ± 0.3066                           & 3.4561 ± 0.4067                          \\
                                     & \textbf{Autoencoder}               & 2.3188 ± 0.1957                           & 1.8579 ± 0.1395                           & 3.5110 ± 0.2027                           & 3.9692 ± 0.3306                           & 6.5044 ± 0.4685                           & 4.7812 ± 0.3819                          \\
                                     & \textbf{EST}                       & -                                         & 1.4472 ± 0.1299                           &                                           & 2.6060 ± 0.1661                           &                                           & 3.4290 ± 0.3368                          \\ \hline
\multirow{4}{*}{\textbf{MLP}}        & \textbf{Baseline}                  & 2.3579 ± 0.2896                           & 1.9006 ± 0.1239                           & 3.8654 ± 0.5083                           & 2.9053 ± 0.1040                           & \textbf{5.1097 ± 0.5556}                           & 3.8222 ± 0.5322                          \\
                                     & \textbf{FFT}                       & \textbf{1.7079 ± 0.0278}                           & 1.6415 ± 0.3577                           & \textbf{2.6119 ± 0.0756}                           & 3.1090 ± 0.2233                           & 5.2028 ± 0.6282                           & 4.1149 ± 0.6749                          \\
                                     & \textbf{Autoencoder}               & 2.4047 ± 0.2135                           & 2.0130 ± 0.1565                           & 4.2544 ± 0.3960                           & 4.1346 ± 0.3691                           & 6.5187 ± 0.5216                           & 5.5296 ± 0.6103                          \\
                                     & \textbf{EST}                       & -                                         & 1.4648 ± 0.2708                           & -                                         & 2.0744 ± 0.1450                           & -                                         & 3.5230 ± 0.3907                          \\ \hline
\multirow{2}{*}{\textbf{RNN + MLP}}  & \textbf{Baseline}                  & 2.2773 ± 0.4501                           & \textbf{0.9288 ± 0.2015}                           & 4.7522 ± 1.1578                           & \textbf{1.2339 ± 0.1775}                           & 4.7450 ± 0.8914                           & 3.2920 ± 1.2359                          \\
                                     & \textbf{EST}                       & -                                         & 1.6028 ± 0.2636                           & -                                         & 1.6164 ± 0.1892                           & -                                         & \textbf{2.6973 ± 0.3968}                          \\ \hline \hline
\end{tabular}
\label{tab:rod_tap_mae_scores}
\end{table*}

\mypara{Robot Tasks.} Our experiments involved three tasks: 
\begin{itemize}
	\item \emph{Tap Localization}, where the robot is tasked to predict the position of a human tap on a held acrylic rod. We tested three rod lengths: 20cm, 30cm and 50cm; 
	\item \emph{Grasp Stability Prediction}, where the goal is to classify whether a human has a stable grasp on an object during robot-to-human handover. We used three different objects, i.e., a rod, a plate, and a box;
	\item \emph{Food Identification}, where the robot has to classify the type of food using a standard fork. There were 7 classes in total: one control (nothing), one bowl of liquid (water), and 5 foods (slices of soft tofu, watermelon, banana, apple, and green pepper).
\end{itemize}
The Tap Localization task was inspired by recent work showing humans can localize contact on held rods~\cite{Miller2018}, and serves as a controlled setting where we could more easily isolate the effect of different factors.
The grasp stability and food identification problems are prototypical real-world tasks that involve tactile signals. Grasp stability prediction may appear an unusual task for dynamic sensing; stability may be better inferred from the static forces. Nonetheless, vibration perception may be useful, e.g., by implicitly inferring the position of the fingers in contact with the object. Prior work on food identification using a sensorized fork~\cite{Bhattacharjee2019} showed that reasonable haptic-based classification is possible. Here, we use a standard dinner fork and dynamic tactile data is sensed using the BioTac hydrophone sensor or the NUSkin.

\mypara{Robot Setup.} For all the tasks above, we used either a BioTac\footnote{We could only use one BioTac due to cost constraints.} or NUSkin sensor, retrofitted onto each finger of a Robotiq 2F-140 gripper attached to a Franka-Emika Panda arm. For the tap localization and grasp stability prediction tasks, we selected a stable position and held the objects in place using position control on the Robotiq fingers. For the food classification task, we used position control to hold the fork at a tilted angle (12 degrees) and programmed the arm to lower the fork at a speed of 1~cm/s. The arm stopped  either when it reached a pre-set height from the table surface, or it detected a Cartesian collision (threshold of 10N/m).

\mypara{Data Collection and Preprocessing.} 
The number of data points varied between tasks: for tap localization, we collected 1000 taps per rod; for grasp stability prediction, 50 grasps were collected per object and per class, and for the food identification task, the robot would skewer each item 50 times. The experiments were performed twice, i.e., once for each sensor. 

Ground truth annotations for the tap localization task were obtained using an Optitrack motion-capture system. 
We varied the tapping direction but kept the tapping angle to be perpendicular to the rod. Care was taken to ensure a good distribution of taps longitudinally along the rod.
Grasp stability was manually annotated; during robot-human handover, a human experimenter would grasp the offered object using stable or unstable grasp. Firm grasps where the object could be supported by the human alone were labelled as stable. Unstable grasps were those where contact was made, but the human's fingers were not in a position to complete the handover without dropping the item. To mitigate damage to the items, they were not actually released by robot.

For the tap localization and grasp stability tasks, we extracted signals from 50~ms before to 250~ms after first contact with the object was detected. For the food identification task, we used sequences starting from 2~s after the arm started moving to the 8~s mark.

\mypara{Vibro-tactile Features.} To evaluate the role of feature learning, we evaluated four different types of features:
\begin{itemize} 
\item \emph{Baseline.} We used the raw PAC signal for the BioTac. For the NUSkin, using the raw spike trains resulted in very poor performance in preliminary experiments. As such, we use binned spike count features instead, i.e., the number of spikes in 5~ms intervals, for the localization and grasp stability tasks. The food identification task used 50~ms bin intervals.  
\item \emph{Fast-Fourier Transform (FFT).} Similar to prior work~\cite{Miller2018}, we used FFT features obtained from the BioTac PAC values and the binned spike trains from the NUSkin. 
\item \emph{Neural Autoencoder.} Autoencoding is a popular unsupervised feature learning method based on a reconstruction loss~\cite{kramer1991nonlinear}. Our encoder comprised a MLP with two hidden layers of 128 and 64 neurons, respectively, that output a 32 dimensional real-vector code. The decoder was also a MLP with two hidden layers, each with 64 and 128 neurons. Both encoder and decoder used ReLU activation units. The reconstruction was performed on the PAC values and spike counts for the BioTac and NUSkin, respectively. 
\item \emph{Event Spike Tensor}~\cite{gehrig2019end}. We generated features by convolving the learnt kernel proposed in \cite{gehrig2019end} on the raw NUSkin spike data. The kernel consists of two fully connected layers, each comprising 30 Leaky ReLU neurons. This convolution resulted in a smoothed sequence, from which we sampled 50 points at equal intervals.
\end{itemize}

\mypara{Machine Learning Methods.} We tested three popular ML methods:
\begin{itemize}
	\item \emph{Support Vector Machine/Regression} (SVM/R), which are popular kernel-based methods based upon the maximum margin principle~\cite{boser1992training,smola2004tutorial}. We experimented with both linear and radial basis function (RBF) kernels;
	\item \emph{Multi-layer perceptron} (MLP), i.e., fully-connected neural networks with two-hidden layers of 16 and 8 neurons with ReLU activations~\cite{nair2010rectified};
	\item \emph{Recurrent Neural Network} (RNN) comprising 16  Gated Recurrent Unit (GRU)~\cite{cho2014learning} cells connected to an MLP with the structure described above. 
\end{itemize}
Each of the features above were fed into a machine learning (ML) model to perform regression (for localization) or classification (for stability or food type prediction). The exceptions were the Autoencoder and FFT features, which were a representation of entire time series and hence, were not used together with the RNN.  

\mypara{Testing Methodology.} We tested each task-sensor-model-feature combination using a standardized setup. We report the average and standard deviation of $K$ scores obtained by repeating the following steps $K$ times: 
\begin{enumerate}
    \item Split the dataset randomly into 90\% training and 10\% testing samples.
    \item Perform grid-search and 4-fold cross-validation on the training samples to obtain model hyperparameters.
    \item Using the hyperparameters, optimize a model on all the training samples.
    \item Test the learned model on the testing samples to obtain a performance score. 
\end{enumerate}
We set $K$ based on available computational resources; $K=5$ for the tap localization task and all RNN models, and $K=20$ for all other combinations.

\begin{table*}[t]
\centering
\caption{\small Accuracy Scores for the Grasp Stability Prediction Task.}
\begin{tabular}{l|l|cc|cc|cc}
\hline \hline
\multirow{2}{*}{\textbf{Methods}}    & \multirow{2}{*}{\textbf{Features}} & \multicolumn{2}{c|}{\textbf{Rod}}       & \multicolumn{2}{c|}{\textbf{Box}}       & \multicolumn{2}{c}{\textbf{Plate}}     \\ \cline{3-8} 
                                     &                                    & \textbf{BioTac (PAC)} & \textbf{NUSkin (2F)} & \textbf{BioTac (PAC)} & \textbf{NUSkin (2F)} & \textbf{BioTac (PAC)} & \textbf{NUSkin (2F)} \\ \hline
\multirow{3}{*}{\textbf{SVM Linear}} & \textbf{Baseline}                  & 0.4792 ± 0.1488       & 0.4850 ± 0.0963 & 0.3792 ± 0.1455       & 0.4900 ± 0.0768 & 0.3583 ± 0.1152       & 0.4750 ± 0.0829 \\
                                     & \textbf{FFT}                       & \textbf{0.6500 ± 0.1225}       & 0.5950 ± 0.1284 & 0.5917 ± 0.1205       & 0.7700 ± 0.1145 & 0.7583 ± 0.1341       & 0.6400 ± 0.1497 \\
                                     & \textbf{EST}                       & -                     & 0.6650 ± 0.1824 & -                     & 0.8950 ± 0.0740 & -                     & 0.7150 ± 0.2104 \\ \hline
\multirow{3}{*}{\textbf{SVM RBF}}    & \textbf{Baseline}                  & 0.6375 ± 0.1376       & 0.6350 ± 0.1314 & 0.5042 ± 0.1192       & 0.4600 ± 0.1020 & 0.4708 ± 0.1850       & 0.4650 ± 0.1492 \\
                                     & \textbf{FFT}                       & 0.6125 ± 0.1520       & 0.6700 ± 0.1229 & 0.5958 ± 0.1187       & 0.8400 ± 0.1241 & 0.6167 ± 0.1275       & 0.6600 ± 0.1562 \\
                                     & \textbf{EST}                       & -                     & 0.7600 ± 0.1715 & -                     & \textbf{0.8900 ± 0.0943} & -                     & \textbf{0.7600 ± 0.1428} \\ \hline
\multirow{3}{*}{\textbf{MLP}}        & \textbf{Baseline}                  & 0.5667 ± 0.1434       & 0.4300 ± 0.1187 & 0.5000 ± 0.1208       & 0.4350 ± 0.1236 & 0.5000 ± 0.1394       & 0.4400 ± 0.1530 \\
                                     & \textbf{FFT}                       & 0.6125 ± 0.1401       & 0.6000 ± 0.1612 & \textbf{0.6333 ± 0.1776}       & 0.8400 ± 0.1319 & \textbf{0.7667 ± 0.1137}       & 0.6150 ± 0.1314 \\
                                     & \textbf{EST}                       & -                     & \textbf{0.7250 ± 0.1479} & -                     & 0.9300 ± 0.0714 & -                     & 0.7450 ± 0.1203 \\ \hline
\multirow{2}{*}{\textbf{RNN + MLP}}  & \textbf{Baseline}                  & 0.5000 ± 0.1394       & 0.6400 ± 0.1497 & 0.6000 ± 0.2261       & 0.5600 ± 0.1497 & 0.5167 ± 0.0624       & 0.4800 ± 0.1720 \\
                                     & \textbf{EST}                       & -                     & 0.4000 ± 0.0894 & -                     & 0.8200 ± 0.2638 & -                     & 0.3200 ± 0.0980 \\ \hline \hline
\end{tabular}
\label{tab:handover_accuracy_scores_v2}
\end{table*}

\section{Results}
\label{ref:results}
In this section, we report our main findings. Briefly, the experimental results support our hypothesis that extended tactile sensing can be achieved with fast vibro-tactile sensing and feature learning. Moreover, using multiple taxels at a high enough sampling rate leads to performance gains. 

\mypara{Can the robot localize contacts using the vibro-tactile signals?} Localization can indeed be performed to a relatively high accuracy; Table \ref{tab:rod_tap_mae_scores} shows that the best models achieved low localization errors of approximately 1~cm, 1.7~cm, and 3.1~cm for the 20, 30, and 50 cm rods, respectively. Among the learned models, the best performance was achieved using the neural models (either the MLP or RNN). That said, good results were also obtained using the SVR (RBF) with FFT features, which we found easier to tune and faster to train. 
The NUSkin generally achieved better performance compared to the BioTac PAC sensor, especially on the 50cm rod where the NUSkin localization was $\approx$1~cm better. 

\mypara{Is vibration perception useful for grasp stability prediction and food identification?} Next, we turn our attention to the ``higher-level'' robot perception tasks.
Accuracies for grasp stability classification (Table \ref{tab:handover_accuracy_scores_v2}) indicate that useful information can be obtained from the dynamic signals; accuracy for the best models was around $60-80\%$ across the three objects. The accuracies were reasonable but could be improved using static tactile sensing and other modalities. The space of stable/unstable grasp configurations is large and possibly, more data is required to obtain better models. The performance across the objects and sensors were mixed; the BioTac (PAC) scores were higher for the plate, but the NUSkin performed better with the box. 

\begin{figure}
\centering
\includegraphics[width=1.0\columnwidth]{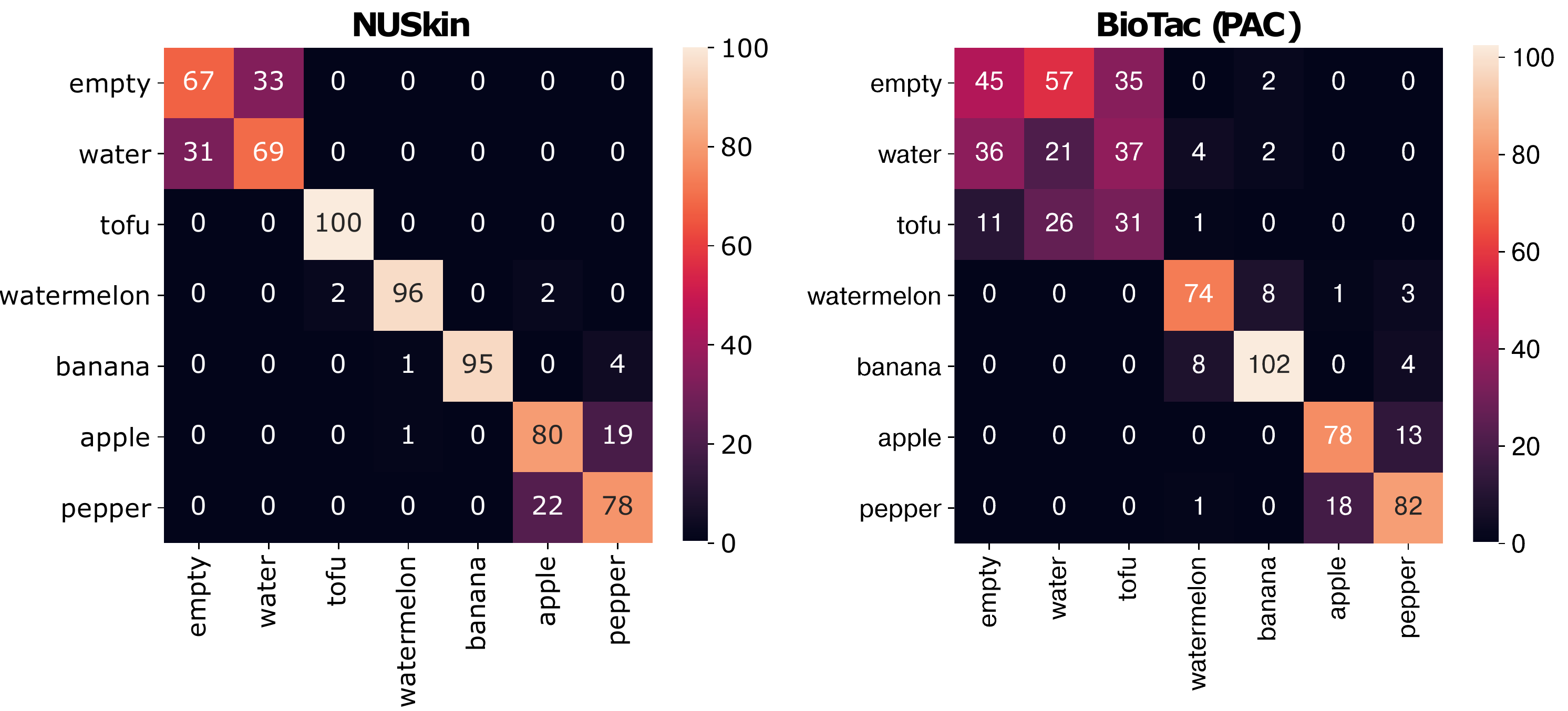}
\caption{\small Confusion Matrices for the NUSkin and BioTac (PAC) on the Food Identification Task. Overall accuracy was 90\% and the errors were clustered around similar objects.}
\label{fig:confusion_matrix} 
\end{figure}

For the food identification task, Table \ref{tab:food_accuracy_scores_v2} shows high accuracies of close to 90\% were obtained using the NUSkin and FFT/EST features.  Fig. \ref{fig:confusion_matrix} shows the confusion matrices for the SVM (RBF) with FFT features obtained from the NUSkin and BioTac. With the NUSkin, the most frequent errors were between the control class (empty plate) and water. Even so, the robot's ability to distinguish between the two classes was surprisingly good ($\approx$ 80\%) given the small forces involved when touching water with a fork. The other errors are reasonable, i.e., between the apple and pepper which had similar consistency, hardness, and texture.

\begin{table}[t]
\centering
\caption{\small Accuracy Scores for the Food Identification Task.}
\begin{tabular}{l|l|c|c}
\hline \hline
\textbf{Methods}                     & \textbf{Features} & \textbf{BioTac (PAC)} & \textbf{NUSkin (2F)} \\ \hline
\multirow{3}{*}{\textbf{SVM Linear}} & \textbf{Baseline} & 0.5129 ± 0.0897       & 0.7586 ± 0.0568      \\
                                     & \textbf{FFT}      & \textbf{0.7486 ± 0.0709}       & 0.8014 ± 0.0708      \\
                                     & \textbf{EST}      &             -           & 0.8671 ± 0.0513      \\ \hline
\multirow{3}{*}{\textbf{SVM RBF}}    & \textbf{Baseline} & 0.5171 ± 0.0711       & 0.7986 ± 0.0499      \\
                                     & \textbf{FFT}      & 0.6186 ± 0.0773       & 0.8357 ± 0.0693      \\
                                     & \textbf{EST}      &                 -       & \textbf{0.8786 ± 0.0578}      \\ \hline
\multirow{3}{*}{\textbf{MLP}}        & \textbf{Baseline} & 0.5114 ± 0.0908       & 0.8000 ± 0.0579      \\
                                     & \textbf{FFT}      & 0.6843 ± 0.0764       & 0.6814 ± 0.1220      \\
                                     & \textbf{EST}      &                   -     & 0.8229 ± 0.0610      \\ \hline
\multirow{2}{*}{\textbf{RNN + MLP}}  & \textbf{Baseline} & 0.3486 ± 0.0492       & 0.3829 ± 0.1602                     \\
                                     & \textbf{EST}      &                    -    & 0.3371 ± 0.0775      \\ \hline \hline
\end{tabular}
\label{tab:food_accuracy_scores_v2}
\end{table}

\begin{figure}
\centering
\includegraphics[width=1.0\columnwidth]{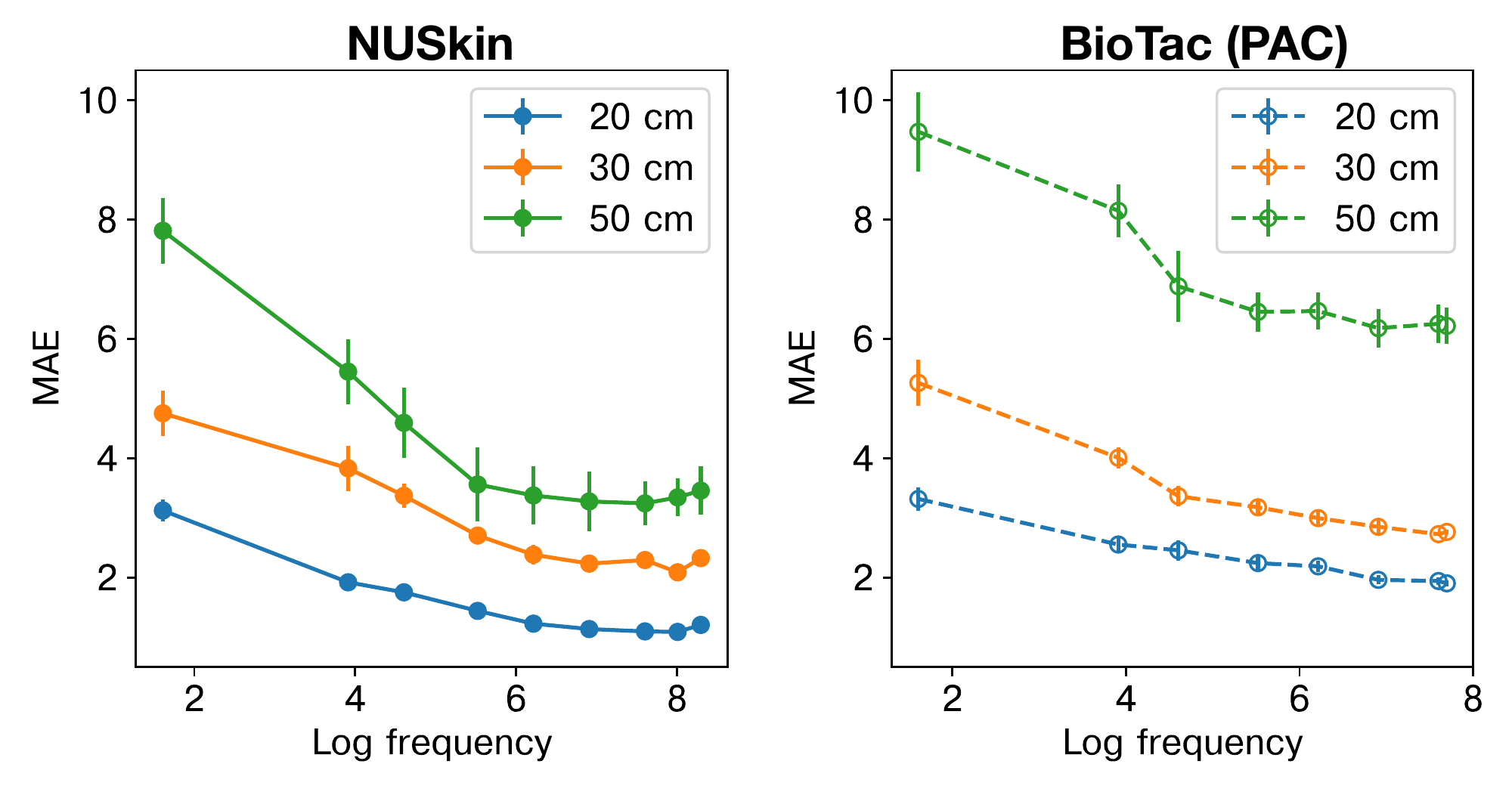}
\caption{\small MAE errors versus the sampling frequency (log-scale) for the NUSkin and BioTac (PAC). Errors fell significantly as the sampling frequency was increased from 5~Hz to 1~kHz, after which improvements were smaller. The NUSkin obtained lower errors in general, with the best performance  in the 2 to 3~kHZ range.}
\label{fig:sampling_results} 
\end{figure}

\mypara{How fast should tactile data be sampled?} To answer this question, we simulated different sampling rates using our obtained data. For the BioTac, we performed standard downsampling via decimation. The NUSkin --- due to its neuromorphic nature --- required a different approach; we iterated through each spike sequence and dropped spikes occurring within a fixed time interval $\tau$ after an ``accepted'' spike. Each taxel was processed independently and the interval $\tau$ was adapted to suit the desired sampling rate. After  downsampling, we applied the same testing methodology described in Sec. \ref{sec:experiments}. 

Fig. \ref{fig:sampling_results} plots localization error for the rod task against the sampling rate when using the SVR (RBF) with FFT features. We see diminishing returns with increasing sampling rate --- error rates fell dramatically as sampling rates increased from 5 to 1000~Hz, after which performance differences were small. The best average performance was obtained in the 2~kHz to 3~kHz range. 
This trend was fairly consistent across all the methods and features. Overall, for the tasks examined, a sampling frequency of 2~kHz was sufficient for good performance.    

\begin{figure}
\centering
\includegraphics[width=0.85\columnwidth]{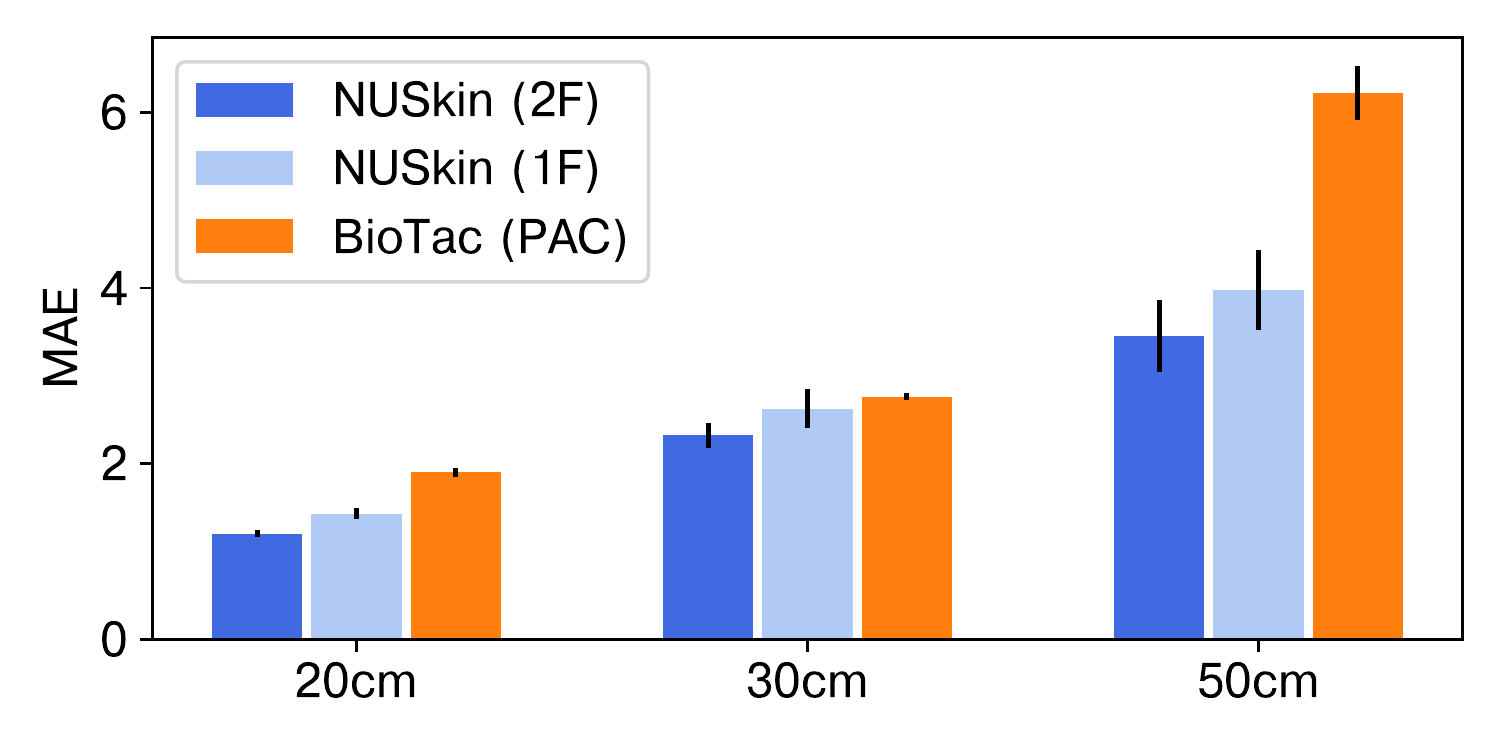}
\caption{\small MAE scores for the NUSkin with 1 or 2 fingers, and the BioTac (PAC). Using two fingers (80-taxels) resulted in lower errors compared to the single finger (40-taxel) version and the BioTac (single hydrophone sensor).}
\label{fig:comparison_half_taxels} 
\end{figure}

\begin{figure}
\centering
\includegraphics[width=0.85\columnwidth]{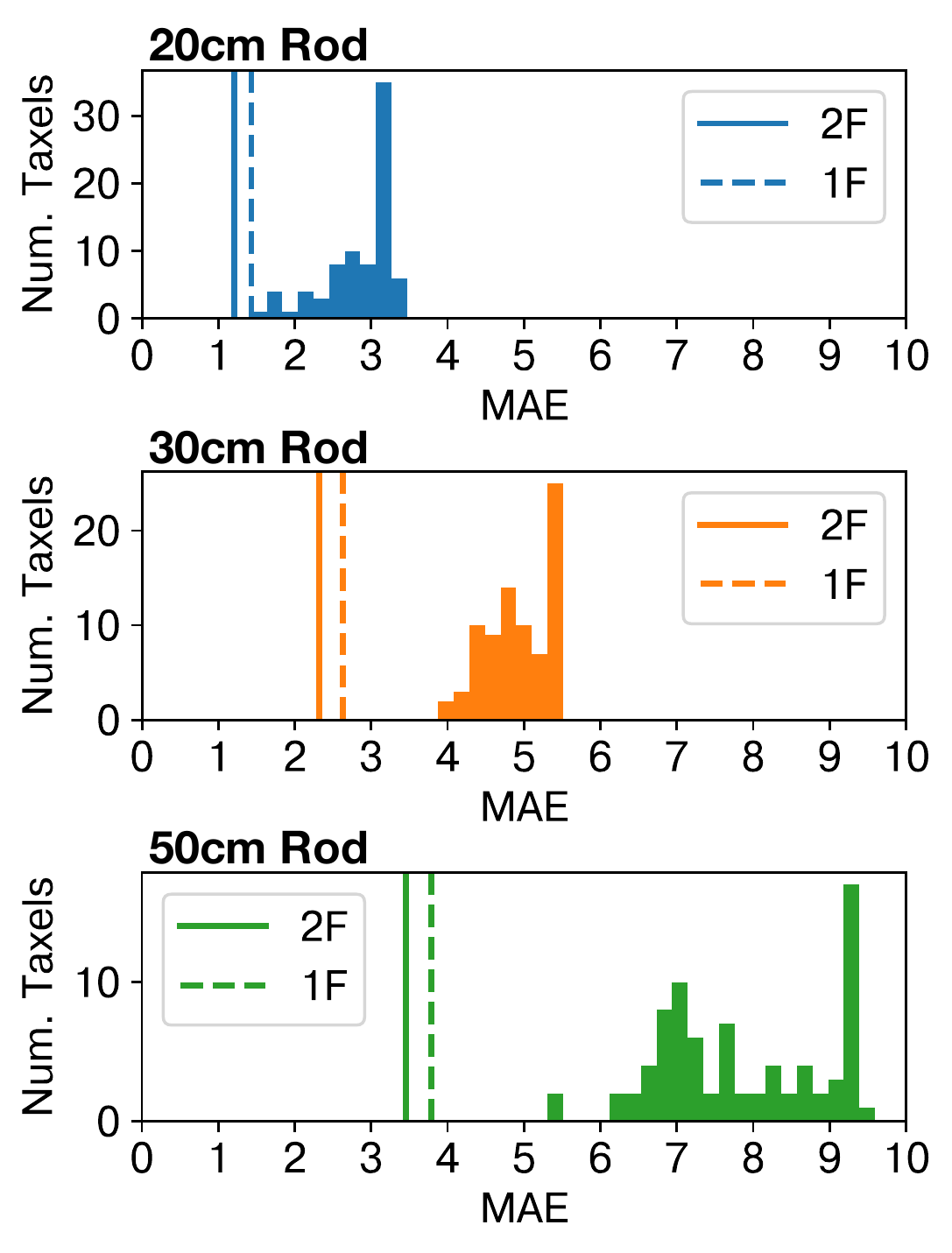}
\caption{\small MAE scores for the single-taxel SVM (RBF) models. Using either 40 taxels (1 finger) or all 80 taxels (2 fingers) resulted in better performance compared to the single-taxel models.}
\label{fig:multi_taxel_results} 
\end{figure}

\mypara{Do multiple taxels lead to better performance?} For this question, we focus on the rod task with the NUSkin sensor. Fig. \ref{fig:comparison_half_taxels} shows the average localization error achieved by the BioTac (PAC) and NUSkin sensors. We further split the NUSkin data into two portions, i.e., the signals gathered by the left and right fingers. The plot shows the performance from using only the left finger (40-taxels), which gave slightly better performance than the right finger in preliminary tests. Overall, we see that using all 80 taxels resulted in better localization. This trend was consistent across the ML methods and features. 

We ran a second test where we trained single-taxel SVMs (RBF kernel); each model was trained using data from a single taxel only. Fig. \ref{fig:multi_taxel_results} shows the histogram plot of the MAEs obtained by the single taxel models, with corresponding scores obtained by the 80-taxel and 40-taxel models marked as vertical lines. Over the three rod lengths, using all 80 taxels (or even just 1 finger) was better than any individual taxel --- this finding suggests the importance of spatially-distributed vibro-tactile sensing.

\section{Discussion and Conclusion}
\label{sec:discussion}

In this work, we showed that tactile perception can be extended beyond the boundary of the robot via vibro-tactile perception. By using appropriate vibratory features, our system is able to accurately localize contacts and classify different contact types.
We hope this work will bring more attention to vibro-tactile sensing and its use in robotics, particularly for tool use. Certainly, there are a variety of unsolved problems to address, e.g.,

\mypara{Non-rigid and multiple-link objects.} In all our experiments, we have worked with rigid objects, which facilitates vibration sensing. We believe extended tactile perception is possible when using non-rigid objects or those with multiple links \emph{if} sufficient information can be transmitted along the object to the tactile sensors. However, the signals may be non-linearly transformed and aliasing may occur, which can hamper performance and require new learning methods. A full investigation with appropriate experiments remains future work.

\mypara{Multi-modal perception.} Vibration sensing is but one mode of sensing, and a coherent representation of the world requires other modalities (e.g., force sensing, vision, auditory). For example, our models for grasp stability prediction on the handover task could potentially be improved by using a combination of static and dynamic tactile sensing, along with multi-modal sequence learning models~\cite{kaiqi21,zhi2020factorized}. 

\section{Acknowledgements}

This work was supported by the SERC, A*STAR, Singapore, through the National Robotics Program under Grant No. W2025d0244. B.C.K. Tee acknowledges grant support from National Research Foundation Fellowship NRFF2017-08.
 
\balance
\renewcommand*{\bibfont}{\footnotesize}
\bibliographystyle{IEEEtranN}
\bibliography{extsense}

\end{document}